\documentclass[conference]{IEEEtran}

\usepackage{cite}
\usepackage{amsmath,amssymb,amsfonts}
\usepackage{algorithm}
\usepackage[noend]{algorithmic}
\usepackage{graphicx}
\usepackage{textcomp}
\usepackage{xcolor}

\DeclareMathOperator{\copt}{copt}
\DeclareMathOperator{\tr}{tr}

\begin{document}

\title{The Convergence of Dynamic Routing between Capsules}

\author{
    \IEEEauthorblockN{1\textsuperscript{st} Daoyuan Ye}
    \IEEEauthorblockA{\textit{School of Informatics} \\
    \textit{Beijing Wuzi University}\\
    Beijing, China \\
    2231108121001Z@bwu.edu.cn}
    \and
    \IEEEauthorblockN{2\textsuperscript{nd} Juntao Li}
    \IEEEauthorblockA{\textit{School of Informatics} \\
    \textit{Beijing Wuzi University}\\
    Beijing, China \\
    Ljtletter@126.com}
    \and
    \IEEEauthorblockN{3\textsuperscript{rd} Yiting Shen}
    \IEEEauthorblockA{\textit{School of Informatics} \\
    \textit{Beijing Wuzi University}\\
    Beijing, China \\
    15648979788@163.com}
}

\maketitle

\begin{abstract}
    Capsule networks(CapsNet) are recently proposed neural network models with new processing layers, specifically for entity representation and discovery of images.
    It is well known that CapsNet have some advantages over traditional neural networks, especially in generalization capability.
    At the same time, some studies report negative experimental results.
    The causes of this contradiction have not been thoroughly analyzed.
    The preliminary experimental results show that the behavior of routing algorithms does not always produce good results as expected, and in most cases, different routing algorithms do not change the classification results, but simply polarize the link strength, especially when they continue to repeat without stopping.
    To realize the true potential of the CapsNet, deep mathematical analysis of the routing algorithms is crucial.
    In CapsNet, each capsule uses the length of the pose vector to represent the probability of the presence of the entity represented by it.
    An irregular nonlinear function is required in the dynamic routing algorithm to keep the vector length less than 1.
    Based on this, Hinton et al. didn’t believe the existence of any sensible objective function that is minimized by the dynamic routing algorithm.
    However, in this paper, we will give the objective function that is minimized by the dynamic routing algorithm, which is a concave function.
    The dynamic routing algorithm can be regarded as nonlinear gradient method to solving an optimization algorithm under linear constraints, and its convergence can be strictly proved mathematically.
    Furthermore, the mathematically rigorous proof of the convergence is given for this class of iterative routing procedures.
    We analyze the relation between the objective function and the constraints solved by the dynamic routing algorithm in detail, and perform the corresponding routing experiment to analyze the effect of our convergence proof.
\end{abstract}


\section{Introduction}
In recent years, convolutional neural networks (CNNs) have had a great impact on the field of deep learning due to their solutions for complex problems and their extensive applications.
CNNs perform very well in image classification for many datasets;
however, they do not explore the spatial relationships between features fully, nor do they classify different variants of the same image.
To address these problems with CNNs, \cite{dyn_rou} and \cite{em_rou} introduced a novel deep learning architecture called capsule network, which was inspired by the mechanism of the visual cortex in biological brains.
The capsule network aims to achieve equivariance to variations, such as pose change, translation, and scaling.
A capsule is made up of a set of neurons that each of them aims to represent an attribute of an object.
The orientation vector of a capsule denotes the pose of an object, while the length of the vector represents the probability of the object’s existence.
To achieve this, the capsule network uses routing algorithms to determine the link strength between each capsule.
The connection strength between the capsules in consecutive layer represents the partial-whole relationship between the objects represented by the capsule.
\cite{em_rou} extends the concept of \cite{dyn_rou} to separate the activation of capsules from their poses.
The agreement between capsules in consecutive layers contributes to strengthening the activation of the higher capsules;
the connection between two capsules becomes fortified as they are activated together, as in the case of the Hebbian rule.
The concept of the expectation-maximization procedure is employed to estimate the connection strength between capsules.

Since the architecture of the capsule network was proposed, many new routing algorithms \cite{can_scr,att_gru} and applications have emerged, including human pose estimation \cite{caps_pose}, emotion recognition \cite{caps_emo}, image classification \cite{caps_hyp_cls,caps_remo,caps_comp_cls}, object segmentation \cite{seg}, medical image segmentation \cite{caps_med,caps_med2,caps_med3,can_scr}, vehicle scene segmentation \cite{caps_veh} and others \cite{caps_app}.
A large number of studies \cite{pt_caps,caps_srv,caps_hyp_cls,caps_comp_cls} have shown that the capsule network has shown more performance than the existing CNN results in many applications.
However, some studies have reported negative results on reinforcement learning \cite{rei_game} and image classification \cite{perf_comp,perf_img_cls}.
This means that we need to study the convergence of routing algorithms and what their real optimization objectives are.

Although many numerical experiments \cite{caps_ms,caps_remo,caps_med} have been carried out on many applied datasets for the capsule network based on routing algorithms, there are few mathematically rigorous studies on the optimization objectives and convergence of routing algorithms.
\cite{dyn_rou} indicate that in order to utilize the length of the pose vector to represent the probability that the paradigm exists in the graph, the vector length must not be greater than one.
This requires the use of non-linear functions, so that the iterative routing process is derived without any meaningful minimization of the objective function.
This iterative process converges in numerical experiments, but lacks strictly mathematical proof.
At the same time, in numerical experiment from \cite{perf_comp}, it is better to iterate 3 times only, as with more iterative steps sometimes the effect is not good.
In \cite{dyn_opt}, part of the routing algorithm proposed by \cite{dyn_rou} as an optimization problem minimizes the combination of the clustering loss and the KL regularization between the current coupling distribution and its final state, and then introduces another simple Routing method.
\cite{imp_rou} performed a detailed analysis of the behavior of the two original routing algorithms: \cite{em_rou} and \cite{dyn_rou}, as well as three other algorithms that were proposed recently: \cite{vis_qa,dyn_opt,grp_equ}.
The experimental results showed that the routing algorithms overly polarize the link strengths, and this issue can be extreme when they continue to repeat without stopping.

In this study, we give the minimized objective function and constraints of the dynamic routing algorithm, and then derive the vector-matrix form of the dynamic routing algorithm according to the nonlinear gradient descent method.
The convergence of dynamic routing algorithm is proved by using the properties of convex function.
This mechanism of analyzing the convergence of dynamic routing can be generalized and applied to a large class of routing algorithms of capsule network \cite{dyn_rou}.
We give the corresponding energy function and get its convergence proof.
By analyzing the objective function and constraints in its optimal problem, it is possible to change the constraints to make the original optimization problem become a series of separable optimization problems, thus reducing the computational complexity.

The rest of this paper is arranged as follows:
in section 2, we will introduce the corresponding mathematical basis, give the nonlinear gradient algorithm and its convergence conditions;
in section 3, we will give the convergence analysis of dynamic routing algorithm and concave objective function;
in section 4, we will perform sufficient experiments to visualize the convergence objectives from section 3 and further pinpont the limitations of this algorithm.
Section 5 gives the corresponding discussion, summary the theories and proof, highlights the experiments conducted, and then gives the conclusion.

\section{Nonlinear Gradient Algorithm and the Convergence}
Considering the convenience of future discussion, we introduce the definition and properties of convex functions firstly, which can be found in \cite{cvx_opt}, so we do not cover the proofs of them.

\subsection{Fundamental Properties of Convex Functions}
There are many definitions of convex functions, and we give the most intuitive definition as following:

\textbf{Definition2.1} For a real-valued function $f(x)$ defined on a convex set $\Omega $ in domain of definition $R^n$ and any two points $x,y \in \Omega $.if $\forall \theta  \in [0,1]$,
\begin{equation}
    (1-\theta) \cdot f(x)+\theta \cdot f(y) \geq f((1-\theta) \cdot x+\theta \cdot y)
\end{equation}
we call f(x) a convex function on the convex set $\Omega$ .

\textbf{Lemma2.1} For a real-valued function $f(x)$ who has a continuity of the first derivative and a first order differential on an open convex set $\Omega $. $f(x)$ is a convex function on $\Omega $ equals to for any two points $x,y \in \Omega $,
\begin{equation}
    f(y) \geq f(x)+\nabla f(x)^{T}(y-x)
\end{equation}
This inequation includes the function value and the gradient value of the two points, so sometimes, it is hard to verify. However, in condition of a second order differential, verification can be solvable.

In order to judge whether a given function $f(x)$ is a convex function, we tend to suppose $f(x)$ a second order differential function. In condition of that, whether $f(x)$ is a convex function depends on whether the Hessian matrix of $f(x)$ is a positive definite matrix.

\textbf{Lemma2.2} For a real-valued function $f(x)$ who has a second-order differential and a continuous second derivative on a convex set $\Omega $.If $\forall x \in \Omega $,the Hessian matrix of this point $H_{f}(x)=\left(\frac{\partial^{2} f}{\partial x_{i} \partial x_{j}}\right)_{n \times n}$ is a real symmetric semidefinite matrix, then $f(x$) will be a convex function

\textbf{Lemma2.3} Suppose $\|\boldsymbol{u}\|=1, \rho \geq -1$, then $I + \rho \cdot u u^{T}$ is positive semi-definite matrix.

\textbf{Theorem2.1}\label{theorem_2_1} Suppose $\psi(z)$ is a doubly differentiable function, and when $z > 0$, $\psi^{\prime}(z) > 0, \psi^{\prime}(z) \geq 0$. Then $\forall x \in R^n$, $f(x) = \psi(\|x\|)$ is a convex function of $x$, where $\|x\|$ is the Euclidean norm of $x$(that is to say,the length of the vector).

\textbf{Proof} Let $g(x) = \|x\|$, from the fundamental properties of norm, $g(x)$ is a convex function. On the basis of \cite{cvx_opt}(page84 Equ(3.11)), then $\psi^{\prime}(z) > 0, \psi^{\prime \prime}(z) \geq 0, f(x) = \psi(\|\boldsymbol{x}\|) = \psi(g(\boldsymbol{x}))$ is a convex function.

For the convenience of the following discussion, except for statements we define the following two functions:
\begin{equation}\label{psidef}
  \psi(z) = z - \arctan (z), \quad z \in R,
\end{equation}
\begin{equation}\label{phidef}
  \varphi(\boldsymbol{x})=\ln \sum_{j=1}^N e^{x_j}, \quad x \in R^N,
\end{equation}

\textbf{Corollary2.1}\label{corollary_2_1} $f(\boldsymbol{x}) = \psi(\|\boldsymbol{x}\|)$ is a convex function, where the $\psi(\cdot)$ is defined in \ref{psidef}.

\textbf{Proof}: when $z \ge 0, \quad \psi^{\prime}(z) = z^2 / (1 + z^2) \ge 0$, $\psi^{\prime \prime}(z) = 2 \cdot z /(1 + z^2)^2 \geq 0 $
The conclusion can be obtained according to \textbf{Theorem 2.1}.

\textbf{Definition2.2} Let $f(x)$ a convex function in $R^n \to (-\infty,\infty]$, the conjugate function of $f(x)$ is defined as:
\begin{equation}
f^*(\xi) = \max_{\mathbf{x} \in R^n}\{\mathbf{x}^T \xi - f(\mathbf{x})\}
\end{equation}

\textbf{Property2.1}\label{property_2_1} Given $f(x)$ a convex function in $R^n$, then its conjugate function $f^{*}$ will also be a convex function.
\\
\textbf{Corollary2.2} As the $\varphi(\cdot)$ defined in Equation (\ref{phidef}), it is easy to verified that $\varphi(x), \quad x \in R^N$ is a convex function.Based on \cite{cvx_opt}, page 93, the following conclusions are true:
\begin{enumerate}
\item $\frac{1}{2} x^T x - \varphi(x)$ is a convex function;
\item $\varphi^*(y)=
    \begin{cases}
        \sum_{i = 1}^M y_i \ln y_i, & \text{if } y_i \geq 0 \\
        & \text{and } \sum_{i = 1}^M y_i = 1 \\
        +\infty, & \text{otherwise}
    \end{cases}$
\item $\varphi^*(y)-\frac{1}{2} y^T y$  is a convex function.
\end{enumerate}

\textbf{Proof}: These properties can be directly verified according to the definition of convex function in a convex set, so we ignore their proof.

Other fundamental definitions and properties can be found in \cite{cvx_opt,imp_conv}

\subsection{Nonlinear Gradient Descent Algorithm}
Consider the following optimization problem:
\begin{equation}
(\copt 1) \min_{x \in \Omega} E(\mathbf{x}), \Omega \subset R^n
\end{equation}
and the gradient descent algorithm will be
\begin{equation}
\boldsymbol{x}(t + 1) = \boldsymbol{x}(t) - \eta_t \cdot \nabla E(\boldsymbol{x}(t))
\end{equation}
where $\nabla E(\boldsymbol{x})$ represents the gradient of $E(x)$ at $x$, $\eta_t$ represents the learning step.
In a general way, big learning steps are always been used at the beginning which will decrease to a small fixed value progressively.
A normal gradient descent algorithm tends to be a unconstrained optimization problem.
However, when it comes to a constrained optimization problem, nonlinear gradient descent algorithm might be a better choice.
Actually, given any differentiable functions $f(x)$, for the optimization problem ($\copt 1$), the nonlinear gradient descent algorithm\cite{imp_conv} will be:
\begin{equation}
    \begin{split}
        \forall \mathbf{u}_0, \text{let } \mathbf{x}_0 = \nabla f(\mathbf{u}_0), \quad t = 0, 1, 2, \ldots \\
        \begin{cases}
            \mathbf{u}_{t + 1} = \mathbf{u}_t - \eta_t \cdot \nabla E(\mathbf{x}_t) \\
          \mathbf{x}_{t + 1} = \nabla f(\mathbf{u}_{t + 1})
        \end{cases}
    \end{split}
\end{equation}
where $f(\mathbf{x})$ is a differentiable function, $\eta_t$ is a positive constant which is relate to $t$(In some cases it can be independent with $t$).
If we choose $f(\boldsymbol{x}) = \boldsymbol{x}^T \boldsymbol{x} / 2$, then the nonlinear gradient descent algorithm turns into the gradient descent algorithm. In order to study the convergence of the nonlinear gradient descent algorithm, we need lemmas as followed:

\textbf{Lemma2.4}\label{lemma_2_4} Let
\begin{equation}
    \begin{split}
        h_{i}(\boldsymbol{x}) &= \boldsymbol{x}^T \mathbf{A}_i \boldsymbol{x}, \quad i = 1, 2 \\
        h(\boldsymbol{x}) &= h_1(\boldsymbol{x}) + h_2(\boldsymbol{x})
    \end{split}
\end{equation}
where $\mathbf{A}_i$ are real symmetric constant matrics.
For real-valued function $F(x)$ and $G(x)$, let $E(x) = F(x) - G(x)$.
If $F(x) - h_1(x)$ and $G(x) - h_2(x)$ are convex functions on the convex set $\Omega$, assume $\exists (u, v) \in \Omega$, $\nabla F(\boldsymbol{v}) = \nabla G(\boldsymbol{u})$,then the following conclusion will be true:
\begin{enumerate}
    \item $E(\boldsymbol{u}) - E(\boldsymbol{v}) \ge h(\boldsymbol{u} - \boldsymbol{v})$
    \item If $A_1 + A_2$ is a positive semi-definite matrix, then $E(\boldsymbol{u}) - E(\boldsymbol{v}) \ge 0$;
    \item If $A_1 + A_2$ is a positive semi-definite matrix, and one of $G(x) - h_2(x)$ or $F(x) - h_1(x)$ will be a strick convex function, then $E(v) < E(u)$ for $u \neq v$
\end{enumerate}

\textbf{Proof}
(Note: $f(x) = x^T A x \Rightarrow f(x) - f(y) - (x - y)^T \nabla f(y) = f(x - y)$)
\begin{enumerate}
    \item As $G(x) - h_2(x)$ is convex function, we can get 
    \begin{equation}
        G(v) - h_2(v) \geq G(u) - h_2(u) + (v - u)^T \nabla (G - h_2)(u)
    \end{equation}

    Therefore
    \begin{equation}\label{equ1}
        G(\boldsymbol{v}) - G(\boldsymbol{u}) \geq h_2(\boldsymbol{v} - \boldsymbol{u}) + (\boldsymbol{v} - \boldsymbol{u})^T \nabla G(\boldsymbol{u})
    \end{equation}

    Similarly,because $\mathrm{F}(\mathbf{x}) - h_{1}(\mathbf{x})$ is a convex function, we can obtain
    \begin{equation}
        F(\boldsymbol{u}) - h_1(\boldsymbol{u}) \geq F(\boldsymbol{v}) - h_1(\boldsymbol{v}) + (\boldsymbol{u} - \boldsymbol{v})^T \nabla (F - h_1)(\boldsymbol{v})
    \end{equation}

    Therefore
    \begin{equation}\label{equ2}
        F(\boldsymbol{u}) - F(\boldsymbol{v}) \geq h_1(\boldsymbol{u} - \boldsymbol{v}) + (\boldsymbol{u} - \boldsymbol{v})^T \nabla F(\boldsymbol{v})
    \end{equation}

    Adding together the left and right sides of the inequality Equation (\ref{equ1}) and Equation (\ref{equ2}) respectively, we can get
    \begin{equation}
        E(\boldsymbol{u}) - E(\boldsymbol{v}) \geq h(\boldsymbol{u} - \boldsymbol{v})
    \end{equation}
    \item If $A_1 + A_2$ is a positive semi-definite matrix, then $h(u - v) \geq 0$, so $E(\boldsymbol{u}) - E(\boldsymbol{v}) \geq h(\boldsymbol{u} - \boldsymbol{v}) \geq 0$
    \item If one of $G(x) - h_2(x)$ and $F(x) - h_1(x)$ is a convex function and $\mathbf{x} \neq \mathbf{y}$, Equation (\ref{equ1}) and Equation (\ref{equ2}) will be strict inequalities, so $E(\mathbf{y}) < E(\mathbf{x})$.
\end{enumerate}

\textbf{Theorem2.2} In nonlinear gradient descent algorithm, if
\begin{equation*}
    \exists h_i(\mathbf{x}) = \mathbf{x}^T \mathbf{A_i} \mathbf{x}, \quad (i = 1, 2)
\end{equation*}
such that $f^*(\mathbf{x}) - h_1(\mathbf{x})$ and $h_2(\boldsymbol{x}) - \eta_t \cdot E(\boldsymbol{x})$ are convex functions, and  $2 \mathrm{A}_1 - \mathrm{A}_2$ is a positive semi-definite matrix, then the following statements are true:
\begin{enumerate}
    \item $E(\boldsymbol{x}_{t + 1}) \leq E(\boldsymbol{x}_t)$
    \item If one of $f^*(\mathbf{x}) - h_1(\mathbf{x})$ and $h_2(\boldsymbol{x}) - \eta_t \cdot E(\boldsymbol{x})$ is a strict convex function or $2 \mathrm{A}_1-\mathrm{A}_2$ is a positive definite matrix, $E(\boldsymbol{x}_{t + 1}) < E(\boldsymbol{x}_t)$
\end{enumerate}

\textbf{Proof} Let $F(x) = f^*(x), \mathrm{G}(\mathrm{x}) = f^*(x) - \eta_t E(x)$, based on the \textbf{Lemma2.4}, the conclusions can be proved.

\section{The Convergence Analysis of Dynamic Routing Algorithm}

We will provide the objective functions to be minimized by dynamic routing procedure and prove that the objective function is a concave function.
Then dynamic routing algorithm will be regarded as an realization of nonlinear gradient descent method.
The final step is to prove that the objective functions is decreasing after each routing step and it will converge to the local optimal solution.

\subsection{The Dynamic Routing Algorithm}

Firstly, we present the whole process of the dynamic routing algorithm and meanings of the vectors,which can be found in \cite{dyn_rou}.
For ease of understanding, we try to use the same symbols to express the same meaning.

\begin{algorithm}
    \caption{The Routing Procedure in \cite{dyn_rou}}\label{algo:dyn_rou}
    \begin{algorithmic}[1]
        \FOR{all capsule $i$ in layer $l$ and capsule $j$ in layer $(l+1)$}
            \STATE $b_ij(0) = 0$
        \ENDFOR
        \FOR{iteration $r = 0, 1, \ldots, K$}
            \FOR{all capsule $i$ in layer $l$}
                \STATE $c_{ij}(r) = \frac{\exp(b_ij(r))}{\sum_{k = 1}^N \exp(b_{ik}(r))}$
            \ENDFOR
            \FOR{all capsule $j$ in layer $(l + 1)$}
                \STATE $s_j(r) = \sum_{i = 1}^M c_{ij}(r) \cdot \hat{u}_{j | i}$
            \ENDFOR
            \FOR{all capsule $j$ in layer $(l + 1)$}
                \STATE $\boldsymbol{v}_j(r) = \frac{\|s_j(r)\|^2}{1 + \|s_j(r)\|^2} \cdot \frac{s_j(r)}{\|s_j(r)\|}$
            \ENDFOR
            \FOR{all capsule $i$ in layer $l$ and $j$ in layer $(l + 1)$}
                \STATE $b_{ij}(r + 1) = b_{ij}(r) + \hat{\boldsymbol{u}}_{j | i}^T \boldsymbol{v}_j(r)$
            \ENDFOR
        \ENDFOR
        \RETURN $v_j(K)$
    \end{algorithmic}
\end{algorithm}

In \cite{em_rou}, the authors guessed that there is no optimization objective function in the dynamic routing algorithm, and \cite{dyn_opt} gave a locally optimizing objective function from a EM algorithm perspective.
They can not give a minimization objective function.
We will propose an objective function in the following and regard the dynamic routing algorithm as the nonlinear gradient descent method to solve this function.
Then we will prove the convergence of the dynamic routing algorithm in a strict mathematic theories based on the convergence of the nonlinear gradient descent method.

\subsection{The Dynamic Routing Algorithm in Matrix Derivative Form}
Parameters in the dynamic routing algorithm are scalar, which is not convenient to analyze its convergence.
We will rewrite them into a vector-matrix form.

In order to simplify the mathematical symbols, we firstly introduce the definition of the derivative of matrix function with respect to matrix.
In fact, for matrix functions, reshaping the matrix into vectors can be direct, but sometimes it is inconvenient to deduce conclusions after the reshaping.
Here we still maintain matrix form.

Usually, we use $C(i, :)$ to represent the $i$-th row of matrix $C$ and $\mathrm{C}(:, j)$ to represent the $j$-th column.

\textbf{Definition3.1} Given matrix $C \in R^{M \times N}$ and a matrix function $f(C) : R^{M \times N} \to R$, then the derivative of $f(C)$ with respect to $C$ is defined by $\nabla_{C} f = \left(\frac{\partial f}{\partial c_{ij}}\right)_{M \times N}$, which will still be a matrix.

If $\textrm{C}$ is a row vector or a column vector, then $C$ can be seen as a matrix whose number of row or column is 1. Therefore we use the same form to represent it.

Let
\begin{equation*}
    \begin{split}
        C = (c_{ij})_{M \times N}, B = (b_{ij})_{M \times N} \\
        i = 1, 2, \ldots, M; j = 1, 2, \ldots, N
    \end{split}
\end{equation*}
where $M$ is the number of capslues in layer $l$ and $N$ is the number of capsules in layer $l + 1$.
Considering the predictive vector of the $j$-th capsule in layer $l + 1$ shares a same dimension, we can concentrate them into a matrix as $\hat{U}_j = (\hat{u}_{j | 1}, \hat{u}_{j | 2}, \ldots, \hat{u}_{j | M})$.

Then the net output of the $j$-th capsule in $l + 1$ layer will be $\boldsymbol{s}_j = \hat{\boldsymbol{U}}_j \boldsymbol{C} \boldsymbol{e}_j$ where $e_j$ represents the $j$-th column of a unit vector whose dimension depends on the context.

The dimension of $\boldsymbol{s}_j$ and neurons in capsule $j$ shares a same number, that is, the dimension of net output of different capsules in $l + 1$ might not be the same. According to abbreviation:
\begin{equation*}
    C(:, j) = C e_j, B(:, j) = B e_j, \quad j = 1,2, \ldots, N
\end{equation*}
is a column vector with M dimention. Then the $b_{ij}(r + 1) \gets b_{ij}(r) + \hat{\boldsymbol{u}}_{ji}^{T} \boldsymbol{v}_j(r)$ in algorithm 1 (presence in Algorithm \ref{algo:dyn_rou}) can be rewritten in a vector form as
\begin{equation*}
    B(:, j)(r + 1) = B(:, j)(r) + \hat{U}_j^{T} v_j(r)
\end{equation*}

Considering that in dynamic routing algorithm (presence in Algorithm \ref{algo:dyn_rou}):
\begin{equation}
    \boldsymbol{v}_j(r) = \frac{\|s_j(r)\|^2}{1 + \|s_j(r)\|^2} \cdot \frac{s_j(r)}{\|s_j(r)\|}
\end{equation}

According to the $\psi(\cdot)$'s definition in Equation (\ref{psidef}), it is easy to verify the following conclusion:
\begin{equation*}
    v_j = \frac{\|s_j\|^2}{1 + \|s_j\|^2} \cdot \frac{s_j}{\|s_j\|} = \nabla_{s_j} \psi(\|s_j\|)
\end{equation*}
Considering $s_j = \sum_{i = 1}^M c_{ij} \cdot \hat{u}_{j | i}$, we can get 
\begin{equation*}
    \hat{U}_j^T v_j = \nabla_{C(:, j)} \psi(\|\hat{U}_j C(:, j)\|)
\end{equation*}

Therefore,the update equation of step 10-11 in Algorithm \ref{algo:dyn_rou}:
\begin{equation*}
    B(:, j)(r + 1) = B(:, j)(r) + \hat{U}_j^T v_j(r)
\end{equation*}
can be replaced by
\begin{equation*}
    \begin{split}
        B(:, j)(r + 1) &= B(:, j)(r) \\
        &+ \nabla_{C(:, j)} \psi(\|\hat{U}_j C(:, j)(r)\|), \\
        &j = 1, 2, \ldots, N
    \end{split}
\end{equation*}
where the $\psi(\cdot)$'s definition is given by Equation (\ref{psidef}).

Furthermore, we definite the function of matrix $C$:
\begin{equation} \label{Psidef}
    \Psi(\boldsymbol{C}) = -\sum_{j = 1}^N \psi(\|\hat{U}_j C(:, j)\|)
\end{equation}
then the equation will be $B(r + 1) = B(r) - \nabla_{C} \Psi(C(r))$.
Considering that $c_{ij} = \frac{\exp(b_{ij})}{\sum_{k} \exp(b_{ik})}$ and the definition of $\varphi(\cdot)$ given in Equation (\ref{phidef}), we can rewrite it in vector form:
\begin{equation}
    C(i, :) = \nabla_{B(i;)} \varphi(B(i, :)), \quad i = 1, 2, \ldots, M
\end{equation}
So the update equation of step 4-5 in Algorithm \ref{algo:dyn_rou} will be:
\begin{equation}
    C(i, :)(r) = \nabla_{B(i, :)} \varphi(B(i, :)(r)), \quad i=1,2, \ldots, M
\end{equation}

Futhermore, we definite the function of matrix $B$ as follows:
\begin{equation}\label{Phidef}
    \Phi(\boldsymbol{B}) = \sum_{i = 1}^M \varphi(\boldsymbol{B}(i, :))
\end{equation}
then the update equation will be:
\begin{equation}
    C(r) = \nabla_{B} \Phi(B(r))
\end{equation}

In conclusion, the dynamic routing algorithm can be a matrix form of nonlinear gradient as followed Algorithm \ref{algo:dyn_mat}.

\begin{algorithm}
    \caption{Dynamic Routing Algorithm by Matrix Forms}\label{algo:dyn_mat}
    \begin{algorithmic}[1]
        \STATE Procedure Routing $(\hat{u}_{j | i}, K, l)$
        \STATE Defining two functions $\Phi(B)$ from (\ref{Phidef}) and $\Psi(\boldsymbol{C})$ from (\ref{Psidef})
        \STATE Initializing $B(0), C(0)$:
        
        $\boldsymbol{B}(0) = 0, \boldsymbol{C}(0) = \nabla_B \Phi(\boldsymbol{B}(0))$
        \FOR{$r = 0, 1, \ldots, K$}
            \STATE $B(r + 1) = B(r) - \nabla_{C} \Psi(C(r))$
            \STATE $C(r + 1) = \nabla_{B} \Phi(B(r + 1))$
        \ENDFOR
        \FOR{all capsule $j$ in layer $l + 1$}
            \STATE $S_j = \sum_i c_{ij} \hat{u}_{j | i}$
            \STATE $v_j = \frac{\|s_{j}\|^2}{1 + \|s_j\|^2} \cdot \frac{S_{j}}{\|s_{j}\|}$
        \ENDFOR
        \RETURN $v_j$
    \end{algorithmic}
\end{algorithm}

\subsection{The Convergence of Dynamic Routing Algorithm}

For further discussion, we need some lemmas as followed:

\textbf{Lemma3.1}\label{lemma_3_1} Assume that matrix $C \in R^{M \times N}$and $f(\boldsymbol{C})=\sum_{j=1}^{N} f_{j}(C( :, j))$ where $f_{j}(\bullet)$ is a M-variable convex function and $f(x)$ is a real-valued function. Then $f(C)$ will be a convex function of $C$;Analogously,if $g(\boldsymbol{C})=\sum_{i=1}^{M} g_{i}(C(i, :))$where$g_{i}(\bullet)$is a N-variable convex function,then $g(\boldsymbol{C})$will be a convex function of $C$.

\textbf{Proof} Because $f(C)$ is a separable function concerning the row vector of matrix, the conclusion is obvious.

\textbf{Corollary3.1} $\Phi(B)$ in Equation (\ref{Phidef}) is a convex function and $\Psi(\boldsymbol{C})$ in Equation (\ref{Psidef}) is a concave function.
It is obvious to deduce the conclusion through \textbf{Corollary 2.1}.

Based on the result of \textbf{Property2.1}, we can get \textbf{Corollary3.2}:

\textbf{Corollary3.2}\label{corollary_3_2} For $\Phi(B)$ in Equation (\ref{Phidef}):
\begin{enumerate}
    \item $\frac{1}{2} \tr(\boldsymbol{B}^T \boldsymbol{B}) - \Phi(\boldsymbol{B})$ is a convex funtion of $G$
    \item $\Phi^*(\boldsymbol{B}) - \frac{1}{2} \tr(\boldsymbol{B}^T \boldsymbol{B})$ is a convex funtion of $B$
\end{enumerate}
where $\tr(\mathrm{A})$ represents the trace of the square matrix.

\textbf{Proof} Based on the definition
\begin{equation}
	\begin{split}
        \frac{1}{2} \tr(\boldsymbol{B}^T \boldsymbol{B}) &- \Phi(\boldsymbol{B}) \\
        &= \sum_{i=1}^M[\frac{1}{2} \boldsymbol{B}(i, :) \boldsymbol{B}(i, :)^T - \phi(\boldsymbol{B}(i, \hat{i}))] \\
        & =\sum_{i = 1}^M f(B(i, :))
	\end{split}
\end{equation}
where $f(\boldsymbol{x}) = \frac{1}{2} \boldsymbol{x}^T \boldsymbol{x} - \varphi(\boldsymbol{x})$ is a convex function
and $\frac{1}{2} \operatorname{tr}\left(\boldsymbol{B}^{T} \boldsymbol{B}\right)-\Phi(\boldsymbol{B})$ is a convex function(based on \textbf{Lemma3.1}).

Analogously
\begin{equation}
    \begin{split}
        \Phi^*(\boldsymbol{B}) &- \frac{1}{2} \tr(\boldsymbol{B}^T \boldsymbol{B}) \\
        &= \sum_{i = 1}^M \left[\varphi^*(\boldsymbol{B}(i, :)) - \frac{1}{2} \boldsymbol{B}(i, :) \boldsymbol{B}(i, :)^T\right] \\
        &= \sum_{i = 1}^M \overline{f}(B(i, :))
    \end{split}
\end{equation}
where $\overline{f}(\boldsymbol{x}) = \varphi^*(\boldsymbol{x}) - \frac{1}{2} \boldsymbol{x}^T \boldsymbol{x}$ is a convex function
and $\Phi^*(\boldsymbol{B}) - \frac{1}{2} \tr(\boldsymbol{B}^{T} \boldsymbol{B})$ is a convex function(based on \textbf{Lemma3.1}).

Based on the above conclusions, we can discuss the convergence of dynamic routing algorithm.

\textbf{Theorem3.1} The $C(r)$ in procedure 2 in dynamic routing algorithm is strictly monotone decreasing concerning the energy function $\Psi(\boldsymbol{C})$, that is, $\Psi(\boldsymbol{C}(r)) - \Psi(\boldsymbol{C}(r + 1)) \geq \|C(r) - C(r + 1)\|_F^2$, where $\|\boldsymbol{C}\|_F$ is the F-norm od matrix $C$.

\textbf{Proof} In \textbf{Lemma2.4}, let
\begin{itemize}
    \item $F(\boldsymbol{C}) = \Phi^*(\boldsymbol{C})$
    \item $G(\boldsymbol{C}) = \Phi^*(\boldsymbol{C}) - \Psi(\boldsymbol{C})$
    \item $h_1(C)=\frac{1}{2} \tr(\boldsymbol{C}^T \boldsymbol{C})$
    \item $h_2(\boldsymbol{C}) = h_1(\boldsymbol{C})$
    \item $h(\boldsymbol{C}) = h_1(\boldsymbol{C}) + h_2(\boldsymbol{C}) = \tr(\boldsymbol{C}^T \boldsymbol{C}) = \|C\|_F^2$
\end{itemize}
We can get $\Psi(\boldsymbol{C}) = F(\boldsymbol{C}) - G(\boldsymbol{C})$.

And based on (1) in Lemma2.1:
\begin{equation}
    \begin{split}
        \Psi(\boldsymbol{C}(r)) - \Psi(\boldsymbol{C}(r + 1)) &\geq \mathrm{h}(C(r)-C(r+1)) \\
        &= \|C(r) - C(r + 1)\|_F^2 \geq 0
    \end{split}
\end{equation}
It follows that $C(r)$ is strictly monotone decreasing concerning the energy function $\Psi(\boldsymbol{C})$, and $\Psi(\boldsymbol{C})$ is bounded in the area constituted between $\mathrm{C} \geq 0$ and $Ce = e$, that is, $\Psi(\boldsymbol{C}(r))$ is convergent.
Based on the basic property of the convergence of discrete dynamical system, we can get the convergence of $\mathrm{C}(r)$.

Combining the matrix form of dynamic routing algorithm and the nonlinear gradient descend algorithm, dynamic routing algorithm can be regarded as one of nonlinear gradient descend algorithm.
Its minimizing objective function is $\psi(C)$ in Equation (\ref{Psidef}) and the constraints  are the following linear inequalities:
\begin{enumerate}
    \item $C_{i,j} \geq 0$
    \item $\sum_{j = 1}^N C_{i,j} = 1, \quad i = 1, \ldots, M$
\end{enumerate}

\textbf{Theorem3.2}\label{theorem_3_2} The constraints optimal problem solved by the dynamic routing is the follows:
\begin{equation}\label{opt2}
    \begin{split}
        min_{\boldsymbol{C}} E(\boldsymbol{C}) &= -\sum_{j = 1}^N \psi(\|\hat{U}_j \boldsymbol{C}(:, j)\|) \\
        \boldsymbol{C}(i, :)e = 1, \boldsymbol{C}(i, j) &\geq 0, \quad i = 1, \ldots, M; j = 1,\ldots, N
    \end{split}
\end{equation}
where $\psi(\cdot)$ in Equation (\ref{psidef}), and $e = (1, 1, \ldots, 1)^T$

\section{Experiments}

\subsection{Experiment Design and Objectives}
To further highlight the significance of our research, we conducted two experiments focuesd on the properties of the dynamic routing algorithm, along with our objective function.
Each experiment was designed to investigate a specific aspect of the algorithm's behavior.

The numerical experiment aimed to evaluate the convergence of the dynamic routing algorithm after each iteration, examining how the algorithm converges as it iterates.
In contrast, the distribution experiment focused on analyzing how the dynamic routing algorithm adjusts the distribution of capsules in subsequent level based on the prediction value of the previous level, with the goal of identifying key characteristics of the algorithm.

\subsection{Numerical Experiment}
The numerical experiment is to verify the convergence of dynamic routing algorithm.
First, the predicted value $\hat{u}_{j|i}$ were randomly generated. Based on the given value, we calculated the value of the coefficient matrix $C$ from Equation \ref{Psidef} for each iteration of routing.
Meanwhile, each column of $C$ were calculated, which represents the prediction value of each capsule of layer $l + 1$, for each iteration.

\begin{figure}[htbp]
    \centerline{\includegraphics[width=0.5\textwidth]{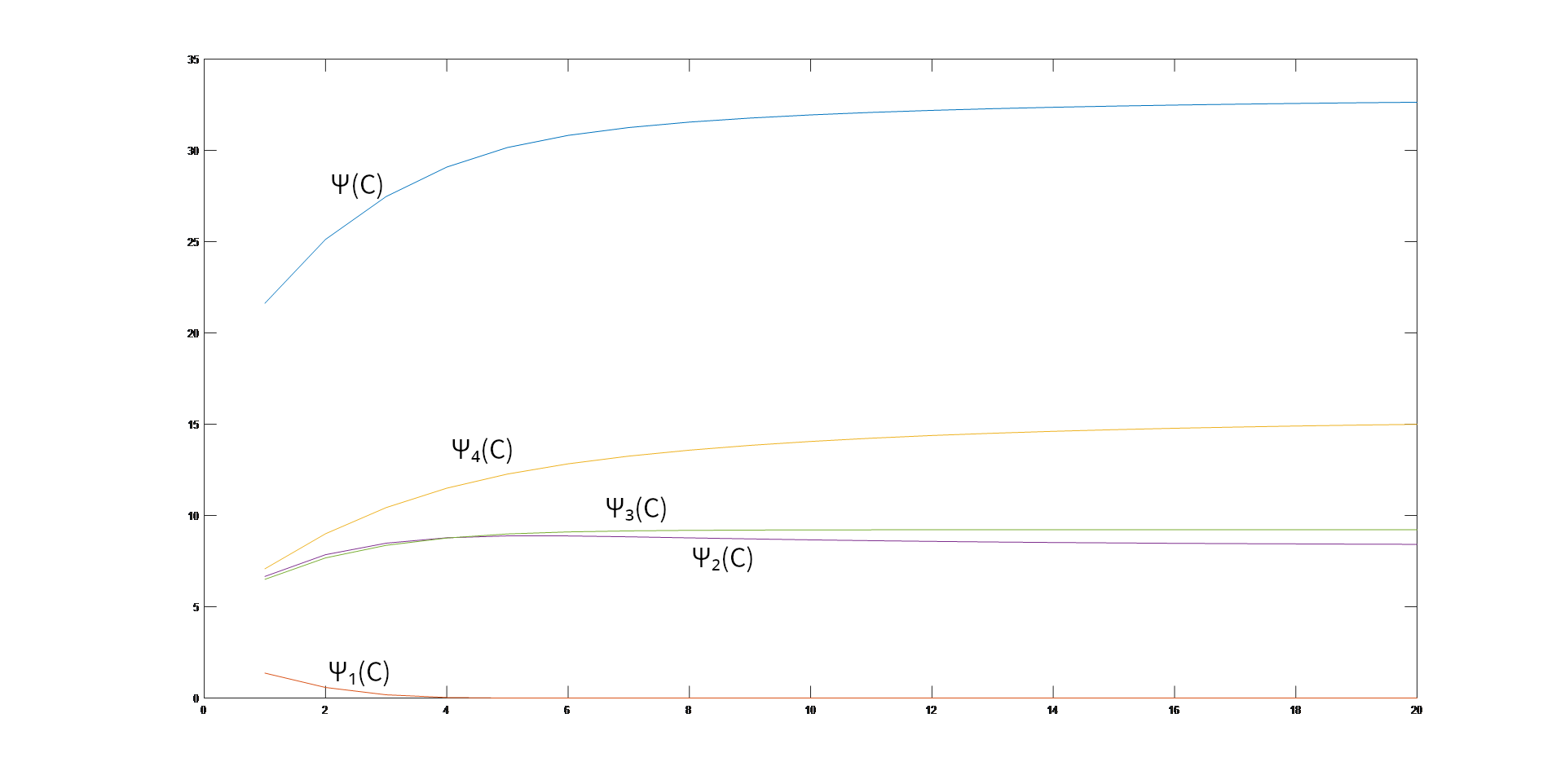}}
    \caption{The values of objective function $\Psi$ (Equation \ref{Psidef}) for all capsules and each capsule during routing process}
    \label{fig:exp1}
\end{figure}

After the calculation of each iteration according to Algorithm \ref{algo:dyn_mat}, we projected the results on a 2D cartesian coordinate plain.
On this plain, the values of $C$ and $C_j$ were projected in y axis, and the iteration count $i$ was projected in x axis, which were demonstrated in Figure \ref{fig:exp1}: a multi-line graph showing the convergence status of matrix $C$ and $C_j$.

In Figure \ref{fig:exp1}, the line above in blue color represents the value of $C$ in each iteration, the lines below in various colors represent the coefficients of different capsules in each iteration. From this figure, $C$ is strictly increasing with iteration, coefficients of several capsules $j$ converged to 0, which indicates that the predictions of these capsules were invalidated during the dynamic routing process.

In summary, this experiment explicitly illustrate the optimization of dynamic routing algorithm by displaying the convergence trend of Equation \ref{Psidef}.
Also, this experiment visualize the optimization effect on each capsules during the routing process.
While some capsules converge to 0, others converge to the proper prediction value, which shows that several capsules are filtered and polarized by the routing process with routing count larger than 5.

\subsection{Distribution Experiment}
The distribution experiment has the same goal as the numerical experiment.
However, the approach is different from numerical experiment.
This experiment directly acquire the input prediction value $u_{j|i}$ of capsules from layer $l + 1$ of each iteration, observe the trend of the distribution afterwards.

\begin{figure}[htbp]
    \centerline{\includegraphics[width=0.48\textwidth]{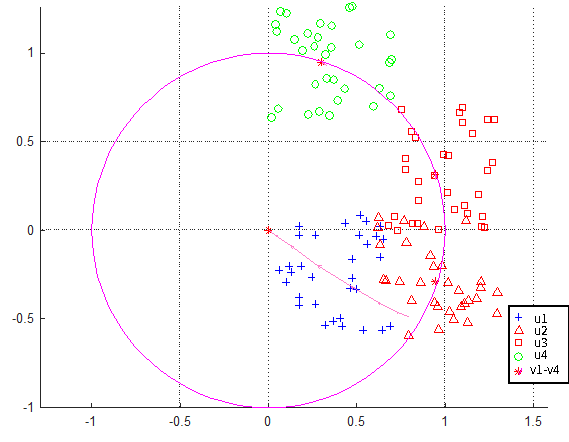}}
    \caption{The distribution map on the input of $l + 1$ layer. Sparse points represent the prediction values of different capsules. The asterisks represent the final output of dynamic algorithm for each capsule.}
    \label{fig:exp2}
\end{figure}

To easily visualize the input, the vector length of capsules in layer $l + 1$ were set to 2, which can be shown in a 2D cartesian coordinate plain. The routing count was set to 20 to get the satisfactory results. The results were shown in Figure \ref{fig:exp2}: a distribution map of various output of each capsules.

In Figure \ref{fig:exp2}, the hollow points distributed around the circle belong to $u_2$ to $u_4$ which gathered around $v_2$ to $v_4$, the blue points distributed around the center of the circle belong to $u_1$ which gathered around $v_1$. In this figure, with the iteration of routing process, the points of $u_1$ is closely distributed around the center, which is $v_1$, as it quickly converge towards the center point, which is (0, 0). The points of $u_2$ to $u_4$ are distributed around $v_2$ to $v_4$, which are almost unit vectors that represents certain features. This indicates that the convergence process filtered the prediction of capsule 1 in layer $l + 1$. It also shows the final orientation during routing process. when iteration reaches 20, $v_2$ to $v_4$ are closely converged to the final point, which shows the orientation and probabilities of capsule $2$ to $4$ that represents the pose each contains.

Ultimately, this experiment not only represents the effect of dynamic routing algorithm that filter certain capsules, but also shows how the unfiltered capsules converge to the optimal orientation/feature during the iteration.

\subsection{Experiment Summary}
By analyzing 2 experiments, certain regularities of the dynamic routing algorithm were discovered.
The convergence of routing algorithm is clarified on the numerical experiment, by observing the increasing and converging trend of the proposed objective function.
In distribution experiment, the relations between coefficient matrix and predictions are exhibited.
It can be observed from the visual output of this experiment that the distribution of the prediction values are scattered around the unit vector of specific orientation.
However, these 2 experiments also indicates that some capsules were reduced to zero, thus "filtered" by the dynamic routing algorithm.
Such polarization problem emerges with the increase of the routing count, which invalidates the utility of certain capsules.

\section{Conclusion}
In this paper, we give the objective function of minimized by the dynamic routing algorithm, and verify that it is a matrix realization form of nonlinear gradient method.
The optimal problem is a linear constrained concave function optimization problem, so it is easy to fall into the Polarization problem.
To further analyze the convergence trend of the algorithm, 2 separate experiments were conducted to visualize the effect on both coefficient and input for each capsules in the next layer, which also highlight the polarization problem.
It is necessary to further analyze the performance of the routing algorithm and seek optimal solution to the polarization problem, so as to propose a practical routing algorithm with a strict mathematical basis.

\bibliographystyle{IEEEtran}
\bibliography{IEEEabrv,refs.bib}

\end{document}